\pdfoutput=1

\documentclass[11pt]{article}

\usepackage{emnlp2021}

\usepackage{times}
\usepackage{latexsym}

\usepackage[T1]{fontenc}

\usepackage[utf8]{inputenc}

\usepackage{microtype}

\usepackage{makecell}
\usepackage{graphicx}
\usepackage{booktabs}
\usepackage{multicol}
\usepackage{multirow}
\usepackage{amsmath}
\usepackage{amssymb}
\usepackage{bbding}
\usepackage{subfigure}
\usepackage{colortbl}
\usepackage{pifont}
\usepackage{bbm}
\usepackage{makecell}
\usepackage{bbding}
\usepackage{mathrsfs}
\usepackage{bm}
\usepackage{xcolor}

\newcommand\ourmethod{\textsc{mt4}CrossOIE}
\newcommand\dataset{\texttt{OpenIE4++}}
\newcommand\mlora{mLoRA}
\newcommand{\xOIE}{CrossOIE}

\title{\ourmethod{}: Multi-stage Tuning for Cross-lingual Open \\ Information Extraction}

\author{
  {\bf Tongliang Li}\textsuperscript{\rm 1}\thanks{\ These authors contributed equally.},
  Zixiang Wang\textsuperscript{\rm 2}\footnotemark[1], 
  Linzheng Chai\textsuperscript{\rm 2},
  Jian Yang\textsuperscript{\rm 2}\thanks{\ Corresponding author.},
  Jiaqi Bai\textsuperscript{\rm 2}, 
  {\bf Yuwei Yin}\textsuperscript{\rm 3}, \\
  {\bf Jiaheng Liu}\textsuperscript{\rm 2}, 
  {\bf Hongcheng Guo}\textsuperscript{\rm 2}, 
  {\bf Liqun Yang}\textsuperscript{\rm 2},
  {\bf Hebboul Zine el-abidine}\textsuperscript{\rm 2},
  {\bf Zhoujun Li}\textsuperscript{\rm 2} \\
  \textsuperscript{\rm 1}Computer School, Beijing Information Science and Technology University; \\ 
  \textsuperscript{\rm 2}Beihang University; \textsuperscript{\rm 3}The University of Hong Kong \\
  tonyliangli@bistu.edu.cn; \\
  \{wangzixiang, challenging, jiaya, bjq\}@buaa.edu.cn; yuweiyin@hku.hk \\ \{liujiaheng, hongchengguo, lqyang, z.hebboul, lizj\}@buaa.edu.cn
}

\begin{document}
\maketitle

\begin{abstract}
Cross-lingual open information extraction aims to extract structured information from raw text across multiple languages. Previous work uses a shared cross-lingual pre-trained model to handle the different languages but underuses the potential of the language-specific representation. In this paper, we propose an effective multi-stage tuning framework called \ourmethod{}, designed for enhancing cross-lingual open information extraction by injecting language-specific knowledge into the shared model. Specifically, the cross-lingual pre-trained model is first tuned in a shared semantic space (e.g., embedding matrix) in the fixed encoder and then other components are optimized in the second stage. After enough training, we freeze the pre-trained model and tune the multiple extra low-rank language-specific modules using mixture-of-LoRAs for model-based cross-lingual transfer. In addition, we leverage two-stage prompting to encourage the large language model (LLM) to annotate the multi-lingual raw data for data-based cross-lingual transfer. The model is trained with multi-lingual objectives on our proposed dataset \texttt{\dataset} by combing the model-based and data-based transfer techniques. Experimental results on various benchmarks emphasize the importance of aggregating multiple plug-in-and-play language-specific modules and demonstrate the effectiveness of \ourmethod{} in cross-lingual OIE\footnote{\url{https://github.com/CSJianYang/Multilingual-Multimodal-NLP}}.
\end{abstract}

\section{Introduction}
Open information extraction (OIE) aims to extract key structured data from an arbitrary domain text in the form of predicates (usually verbals or verbal phrases) and their corresponding arguments~\cite{survey_oie}, without pre-defined relation schemas. Considering the sentence (\textit{``Joe Biden became the US president in the year 2021''}), three tuples are expected to extract by OIE systems: (\textit{Joe Biden; became; the US president}), (\textit{Joe Biden; became the US president; in the year 2021}) and (\textit{Joe Biden; became; the US president; in the year 2021}). Due to the domain independence and scalability, OIE provides powerful help for downstream tasks like question answering~\cite{OIE_downstream_apps, OIE_qa}, summarization~\cite{OIE_summarization}, and knowledge graph completion~\cite{OIE_kgs}.

Most of the existing methods are highly dependent on the labeled data and do not perform well in low-resource languages. Multi2OIE~\cite{multiligual_OIE} is the first neural-based method to tackle OIE task in multiple languages and achieves a satisfactory performance on cross-lingual transfer. 
Large language models~\cite{gpt3,gpt4,llama2} have exhibited extraordinary abilities and have been widely applied to various tasks, including OIE~\cite{inpars,gpt_re}, and other tasks~\cite{lin2023evolutionary,yin2023finpt}.
Despite the success of the existing advances in OIE, the following limitations have not been fully investigated yet: (1) Fine-tuning the entire language model may result in its previously learned knowledge being forgotten due to the catastrophic forgetting \cite{pet_head}. (2) Another limitation is the lack of robustness in handling low-resource languages and can not tackle different languages simultaneously \cite{dozens_mnmt}. 

In this paper, we propose a multi-stage tuning framework for CrossOIE to encourage knowledge sharing among different languages. Inspired by the previous work~\cite{deberta,disentangling_position_information}, the word embedding of the cross-lingual pre-trained model is tuned to align the representations of different languages by freezing the encoder in the first stage while other components are adjusted in the second stage. Given the strong OIE model, we add the mixture-of-LoRAs (\mlora{}) into the fixed model for different languages, where all languages depend on the same backbone model, and adjust the output by combing different low-rank adapters. Besides, we leverage the large language model (LLM) as the cross-lingual annotator to label the multi-lingual raw corpora originating from the English data. Finally, we combine model-based and data-based transfer in our framework to improve the performance of the cross-lingual OIE.

Our contributions are summarized as follows: 
\begin{itemize}
    \item We propose a multi-stage tuning framework called \ourmethod{} for CrossOIE that combines model-based and data-based methods to transfer knowledge into the pre-trained model, which has the superiority in extracting the tuples across different languages. 

    \item We build a new multi-lingual corpus called \texttt{\dataset}, which consists of the original English data and their counterparts of other languages via the large language model with the cross-lingual prompt.

    \item We conduct extensive experiments on multiple languages (English, Arabic, Chinese, German, Spanish, and Portuguese) and the results demonstrate that \ourmethod{} outperforms baseline models on most languages of different benchmarks. Finally, we perform an extensive analysis and reveal the nature of OIE in different languages.
\end{itemize}

\section{Cross-lingual OIE}
Given the source information extraction model $\Theta_{IE}^{src}$ only trained on the source information extraction dataset and the target raw sentence $x=(x_1,\dots,x_m)$ with $m$ words, the zero-shot cross-lingual information extraction aims to identify potential arguments and focuses on extracting predicates among different arguments mentioned in the raw text. Then, we can obtain a list of tuples $T=\{T_1,\dots,T_N\}$, where $T_i=(a_i^1,p_i,a_i^2,\dots,a_i^q)$ is the $i$-th tuple, $p_i$ denotes the predicate in $T_i$ and $a_i^j$ is the $j$-th argument of $p_i$. The $a_i^1$ is considered as the subject and $a_i^2,\dots,a_i^q$ are objects associated with $T_i$. The problem definition of zero-shot cross-lingual open information extraction (OIE) is described as:
\begin{MiddleEquation}
\begin{align}
    P(T|X) = \prod_{i=1}^{N}P(T_i|x;\Theta_{IE}^{src})
    \label{problem_definition}
\end{align}
\end{MiddleEquation}where the tuples $T$ are derived from the target raw sentence $x$. $T_i$ is the $i$-th tuple. The source language has annotated labels but the target corpora have no accessible handcrafted labels. $P(T|x)$ represents the predicted distributions of labels. The source information extraction model $\Theta_{IE}^{src}$ trained on the source annotated corpus is expected to be evaluated on the target language without any labeled dataset. Multi2OIE \cite{multiligual_OIE} aims to extract facts in a sentence without relying on pre-defined schemas. It aims to be more flexible and capable of handling a wider range of textual input. In this work, we propose to unify the model-based transfer from the cross-lingual pre-trained model and data-based transfer with machine translation to transfer knowledge from the source language to the target language.

\begin{figure*}[ht]
\begin{center}
\includegraphics[width=1.0\textwidth]{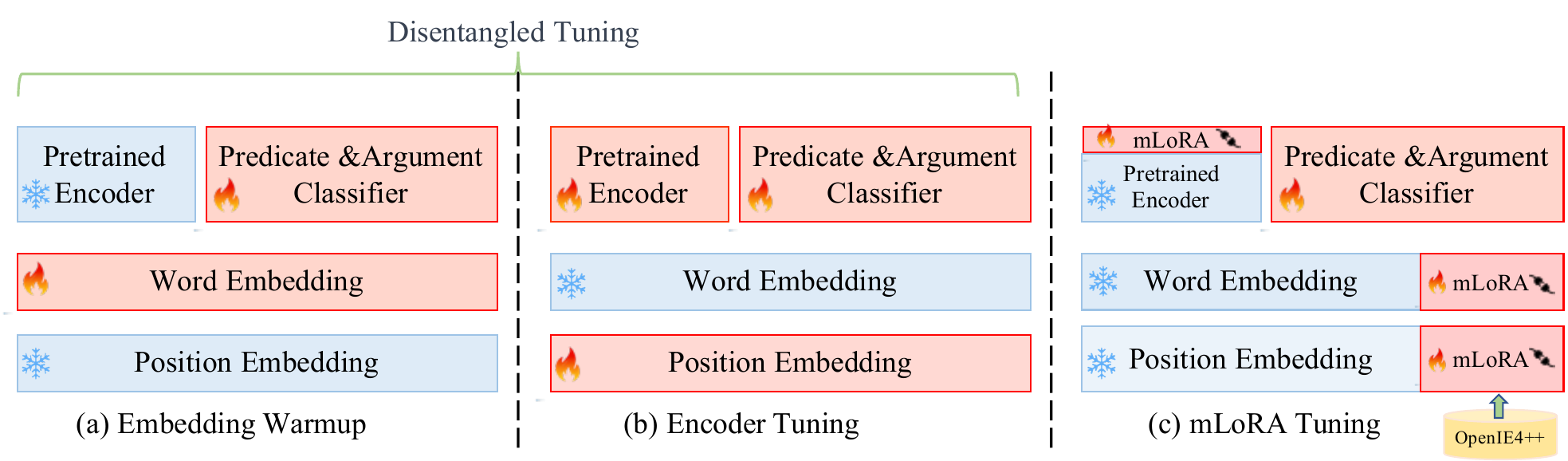}
\caption{The training sketch of \ourmethod{}, where the blue ice icon indicates parameter-frozen modules while the red fire icon denotes trainable ones. In the first stage (a) and second stage (b), we align the semantic representation by disentangled tuning. In the third stage (c), we introduce the mixture-of-LoRAs to compose the language-specific representations for prediction. Additionally, we construct the multi-lingual corpora \dataset to trigger the cross-lingual generalization ability.}
\label{multi_stage}
\end{center}
\end{figure*}

\begin{figure}[t]
\begin{center}
\includegraphics[width=1.0 \columnwidth]{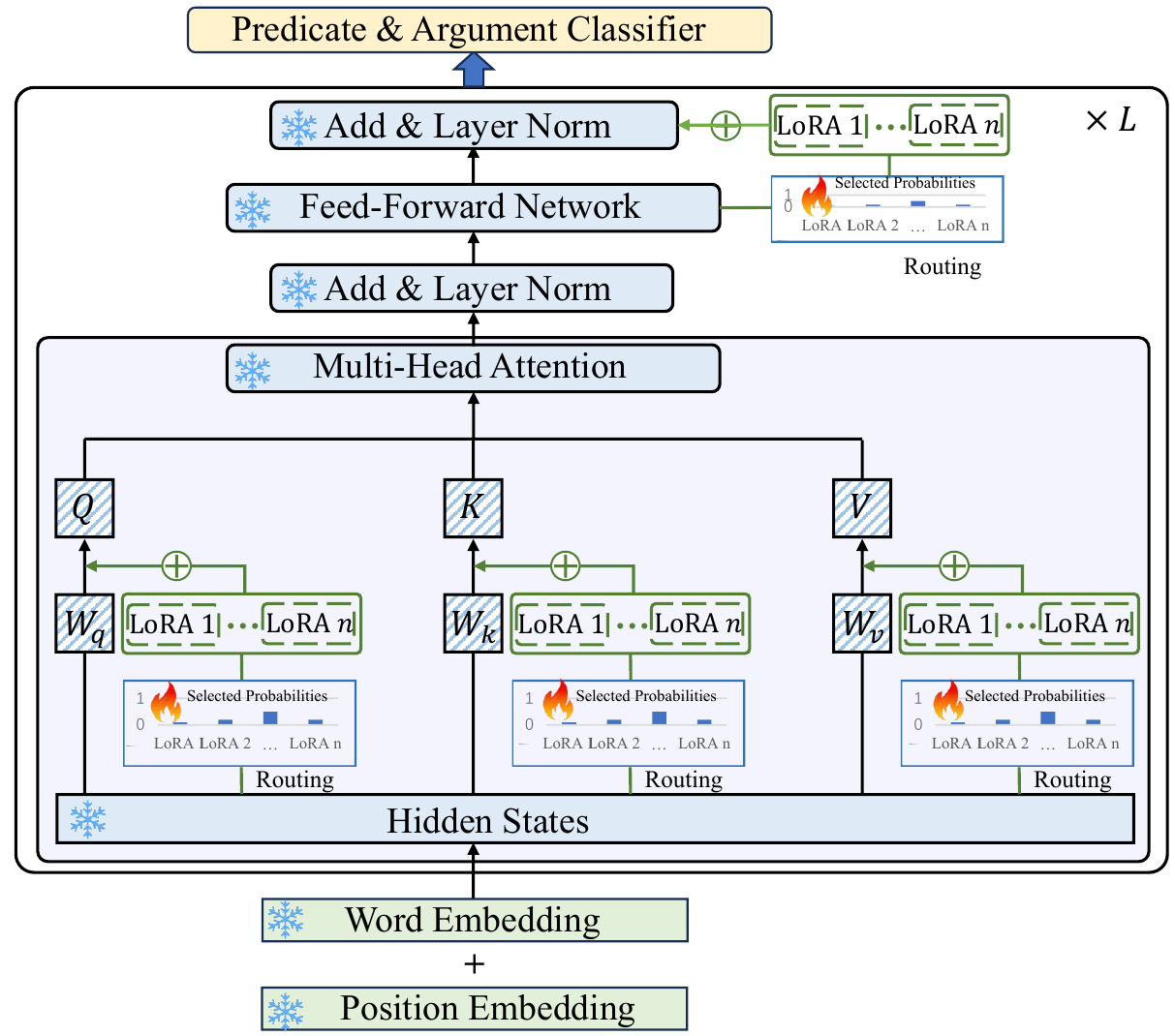}
\caption{The overview of \ourmethod{}. We calculate the selection probabilities of all LoRA adapters and choose the top-$k$ LoRA experts obeying the probability values. Selection probabilities are determined by the hidden state of each layer.}
\label{mlora}
\end{center}
\end{figure}
\section{Methodology}
    In this section, we propose the multi-stage fine-tuning method for cross-lingual OIE as shown in Figure \ref{multi_stage}, where we align the semantic representation in the first stage and adjust other model components in the second stage by disentangled tuning. Next, we introduce the mixture-of-LoRAs to compose the language-specific representations for prediction. Furthermore, we trigger the cross-lingual generalization ability of a large language model using the cross-lingual prompt to construct the multi-lingual corpora \dataset to further augment the cross-lingual transfer.

\subsection{Backbone Model}
Given the input sentence $x=\{x_1,\dots,x_n\}$ of language $L_{k}$, our backbone model first predicts the predicate tagset $t^{p}=\{t_1^{p},\dots,t_n^{p}\}$ with a predicate head, and then outputs the argument tagset $t^{a}=\{t_1^{a},\dots,t_1^{a}\}$ with an argument head. Following the previous work \cite{multiligual_OIE}, we use BIO (Beginning-Inside-Outside) sequence-labeling scheme to tag the predicates and arguments in a sentence. The backbone is a two-step $n$-ary extraction which is first extracting all predicates and then the arguments associated:
\begin{align}
\begin{split}
P(T^{p},T^{a}|x) = P(T^{p}|x)P(T^{a}|x,T^{p})
\end{split}
\end{align}where $T^{p}$ and $T^{a}$ have the same length and OIE is regarded as an $n$-ary extraction task. 
\paragraph{Multi-head Attention}
Given the input embedding $X$, we project $X$ into $Q$ as the query, $K$ as the key, and $V$ as the value in the self-attention module to extract the representations:
\begin{align}
\begin{split}
X_{attn} = \overset{H}{\underset{h=1}{\big\|}}  \mathtt{SF} \left (\frac{QK^T}{\sqrt{d_k}}\right)V
\label{self_attention}
\end{split}
\end{align}where $\texttt{SF}(\cdot)$ denotes the softmax function, and $\|_{h=1}^H$ is the feature concatenation of the $H$ attention heads. The input $X$ is projected into $Q=W^qX,K=W^kX,V=W^vX$ with the learned matrix $W^{q},W^{k},W^{v}$. After the self-attention module, other standard operations (e.g. feed-forward network) are used. Finally, we obtain the representations $H=\{h_1,\dots,h_n\}$ and $H \in \mathbb{R}^{n \times d_k}$

\paragraph{Predicate and Argument Extraction}
The sentence representations $H$ are then fed into a predicate prediction head that consists of a feed-forward network and a softmax layer to classify each token into a predefined predicate tag. We obtain the predicted tags $t^p=\{t^1,\dots,t^n\}$ and the cross-entropy loss $\mathcal{L}_p$ is optimized for predicate extraction.

After predicting the predicate tags, we sum the average representation of the predicted predicate $h^p$ and each word representation $H$, which are then fed into an argument extractor comprised of $N_{2}$ multi-head attention blocks as in Equation \ref{self_attention}. Finally,
the output of the multi-head attention block is fed into the argument classifier. 

\subsection{Disentangled Tuning}
Let $\Theta=\{\theta_{p},\theta_{w},\theta_{b},\theta_{c}\}$ denote all model parameters, where $\theta_{p}$ is the position embedding, $\theta_{w}$ is the word embedding, $\theta_{b}$ is the pre-trained model, and $\theta_{c}$ is the predicate and argument classifier. For a token at position $i$ in a sequence, we represent it using two vectors, $H_i$ and $P_i$, which represent its word and position embedding, respectively. The calculation of the cross-attention score $A_{i,j}=\frac{(H_i + P_i)(H_j + P_j)^T}{\sqrt{d}}$ between tokens $i$ and $j$ can nearly be decomposed into four parts as (omitting scaling factor $\sqrt{d}$):
\begin{align}
\begin{split}
A_{i,j} = \rlap{ $ \overbrace{\phantom{H_iH_j^{T}+H_iP^T_{j} + H_jP^T_{i}}}^{\text{(1) Content-based Terms}} $}H_iH_j^{T}+\underbrace{H_iP^T_{j} + H_jP^T_{i} + P_{i}P^T_{j}}_{\text{(2) Position-based Terms}}
\end{split}
\end{align}where $A_{i, j}$ denotes the attention score between position $i$ and $j$. The content-based term (1) $H_iH_j^{T}+H_iP^T_{j} + H_jP^T_{i}$ optimizes the word embedding while the position-based term (2) $H_iP^T_{j} + H_jP^T_{i} + P_{i}P^T_{j}$ relates to the position embedding.

Zero-shot inference depends on the cross-lingual generalizability of the pre-trained model to conditions unseen in training. In the context of zero-shot OIE, the input should ideally be encoded into a language-agnostic representation. Inspired by the previous work \cite{deberta,disentangling_position_information}, we propose a disentangled tuning strategy to relax the constraint between the word and position information. As shown in Figure \ref{multi_stage}, we only tune the content-based term in the first stage by only tuning the word embedding parameters ($\theta_{w}$) and classifier ($\theta_{}$):
\begin{align}
\begin{split}
\mathcal{L}^{(1)} = - \mathbb{E}[\log P(T|x;\Theta_1=\{\theta_{w},\theta_{c}\})]
\end{split}
\end{align}where $x$ is the input sentence and $T$ are the extracted tuples.

Then, other components continued to be tuned $\Theta_2=\{\theta_{p},\theta_{b},\theta_{c}\}$ to optimize the position-based term (2)  $H_iP^T_{j} + H_jP^T_{i} + P_{i}P^T_{j}$ by freezing the word embedding matrix:
\begin{align}
\begin{split}
\mathcal{L}^{(2)} = - \mathbb{E}[\log P(T|x;\Theta_2=\{\theta_{p},\theta_{b},\theta_{c}\})]
\end{split}
\end{align}where $x$ is the input sentence and $T$ are the extracted tuples.

\subsection{Mixture-of-LoRAs for \xOIE{}}
LoRA \cite{lora} is a tuning technique in LLMs, which enables efficient and flexible transfer by introducing task-specific modifications to a fixed pre-trained model. For the cross-lingual transfer, we use the mixture-of-LoRAs (\mlora{}) for different languages, where a group of LoRA adapters is lightweight compared to the pre-trained model. The adapters with a low-rank down-project matrix and up-project matrix can be directly inserted into the pre-trained embedding, attention, and feed-forward network. Given the source sentence $x=\{x_1,\dots,x_n\}$ of $n$ tokens and a group of $T$ LoRA experts, we use \mlora{} to learn the language-sensitive representations for the same task of different languages:
\begin{align}
\begin{split}
    h_{a}^{L_i} = \mathcal{A}_{\theta_{g(L_i)}}(h^{L_i})
    \label{lora}
\end{split}
\end{align}where $g(L_i)$ are selected LoRA experts derived from the language representations. $\mathcal{A}(\cdot)$ denotes the LoRA adapter module and $\theta=\{\theta_1,\dots,\theta_{T}\}$ denotes the adapter pool. $\mathcal{A}_{\theta_{g(L_i)}}$ is calculated by:
\begin{align}
\begin{split}
    \mathcal{A}_{\theta_{g(L_i)}}(h^{L_i}) = h^{L_i} + \sum_{A_t,B_t \in \mathcal{S}(e)} \alpha \Delta W h^{L_i}
    \label{lora_def}
\end{split}
\end{align}where $\Delta W = BA$ is denoted by a low-rank decomposition ($A \in \mathbb{R}^{d \times r} \land B \in \mathbb{R}^{r \times d} \land r \ll d$). The matrices $A$ and $B$ are initialized by a random Gaussian distribution and zero. $\alpha$ is the scaling factor and $r$ is the inner dimension. $S(e)$ denotes the subset from the 

In Equation \ref{lora_def}, all experts only require fine-tuning a small number of language-specific parameters instead of all parameters of the pre-trained model. Thus, we can simultaneously train multiple experts for different languages, which all share the same freezing pre-trained parameters. We use multiple adapters from the selected subset to maximize the transfer of knowledge across languages:
\begin{align}
\begin{split}
    g(L_i) = \mathop{\texttt{TopK}}\left(\frac{exp(\alpha^{L_i}_{j})}{\sum_{t=1}^{T}exp(\alpha^{L_i}_{t})}\right)
\end{split}
\end{align}where $\texttt{TopK}(\cdot)$ is the selection function, where we calculate the selection probabilities of all LoRA adapters and choose the top-$k$ LoRA experts obeying the probability distribution. $\alpha^{L_i}_j$ is a scalar from the representations of language $L_i$ (We use the hidden state of the special token \texttt{[CLS]} of each layer). $S(e)=\{(A_k,B_k)\}_{k=1}^{K}$ and $\alpha_{j}^{L_i}$ is used to incorporate the different experts.


We project the language representation $e^{L_i}$ of language $L_i$ into the LoRA expert distribution using the learned matrix $W_a \in \mathbb{R}^{d \times T}$, where $d$ is the hidden size and $T$ is the number of experts. The weight of LoRA expert $\alpha_j^{L_i}$ is calculated by:
\begin{align}
\begin{split}
    \alpha^{L_i} = e^{L_i}W_a
\end{split}
\end{align}where $\alpha=\{\alpha_1,\dots,\alpha_T\}$. For all modules of the pre-trained model, we leverage the mixture-of-LoRAs to learn the language-sensitive representations for the different input sentences by activating top-$k$ experts.

\subsection{Multi-lingual Training} \label{sec:multi-lingual_training}
\paragraph{LLM as Cross-lingual Annotator} Large language models (LLMs) equipped with a growing arsenal of prompt-based methods offer the powerful off-the-shelf few-shot capability to the cross-lingual NLP task. To facilitate the generalizability of the cross-lingual model, we design the cross-lingual prompt $P=\{p_1,p_2\}$ to trigger the potential of LLM. The cross-lingual annotation problem can be decomposed into translation procedure and OIE annotation. We use the prompt $p_1$ for translation and the prompt $p_2$ for OIE annotation as:
\begin{align}
\begin{split}
    P(y,T|X) = P(y|x,p_1)P(T|y,p_2)
    \label{generation}
\end{split}
\end{align}where $y$ is the target translation of $X$ and $T$ is the corresponding extracted tuples. The source sentence is translated into the target sentence with the first prompt $p_1$ and then extracted into multiple tuples with the second prompt. Table \ref{prompt} shows the detailed chain-of-thought prompt for cross-lingual annotation using the large language model.
\paragraph{Multi-lingual Training Objective} Given the supervised corpus $D_{L_i}$, we expand the source corpus to the multi-lingual corpora $D=\{D_{L_1},\dots,D_{L_{K}}\}$ of $K$ language using LLM. The training objective of the \xOIE{} can be described as:
\begin{align}
\begin{split}
    \mathcal{L}_{m} = -\frac{1}{K} \sum_{i=1}^{K} \mathbb{E}_{x,T \in D_{L_i}}\log(T|x)
    \label{multi-lingual_training}
\end{split}
\end{align}where $x$ and $T$ are input sentences and extracted tuples from the multi-lingual corpora.

\begin{table*}[t]
\centering
\resizebox{1.0\textwidth}{!}{
\begin{tabular}{c|l}
\toprule
Task      &  Prompt     \\ \midrule
$p_1$: Translate \texttt{[X]} to \texttt{[Y]}  & \makecell[l]{You are a \textcolor{red}{translator}. Please translate the following English text into the \texttt{[L]}: \texttt{[X]}}                    \\ \midrule 
\makecell[c]{$p_2$: Annotate \texttt{[Y]}}          &  \makecell[l]{You are an \textcolor{red}{Information Extraction expert}. The following are the extraction results \\ of \texttt{[Y1]}, which are represented by Subject, Relation, and Object: \texttt{[S1]}, \texttt{[R1]}, \\ \texttt{[O1]} 
Please refer to the extraction results above, extracting a triple that corresponds \\ Subject, Relation, and Object from the translated sentence: \texttt{[Y]}. Note that the subject, \\ relation, and object must originate from the continuous segment of the sentence. The \\ output format must be the same as the sample above.}           \\ 

\bottomrule
\end{tabular}
}
\caption{Prompts and their usage for the large language model (\texttt{gpt-3.5-turbo}). \texttt{[X]} is the source sentence of English and \texttt{[Y]} is the translated sentence of,target language \texttt{[L]}. \texttt{[S1]},\texttt{[R1]}, and \texttt{[O1]} denote the subject, relation, and object of the target translated sentence: \texttt{[Y1]}, where we provide the example consisting of the target sentence \texttt{[Y1]} and its extraction results  (\texttt{[S1]},\texttt{[R1]},\texttt{[O1]}) for the few-shot extraction.}
\label{prompt}
\end{table*}

\begin{table}[htb]
\centering
\resizebox{1.0\columnwidth}{!}{
\begin{tabular}{l|ccccc}
\toprule
 Statistics & Ar & De  & Es & Pt  &  Zh \\
\midrule
      

  \#Sent.     & 2,000 & 10,000 & 10,000 & 10,000 & 10,000 \\
  \#Tuples  & 3,091 & 15,339 & 16,842 & 17,033 & 15,871  \\
  Max\_len  &    52 &     63 &     68 &     75 &   110  \\
  Min\_len  &     4 &      4 &      4 &      4 &     5  \\
  Avg\_len & 19.9  &   21.9 & 24.8   &   24.0 &   35.4  \\
\bottomrule
\end{tabular}}
\caption{Statistics of sentence number, tuple number, maximum sentence length, minimum sentence length, and average sentence length in \texttt{OpenIE4++}.}
\vspace{-10pt}
\label{tab:dataset_statistics}
\end{table}

\section{Experiments}
\subsection{Experimental Setup}
\paragraph{Datasets}

\begin{itemize}
\item Our training data is the same as that used in~\cite{multiligual_OIE, SpanOIE} for the first and second stages.
This English training dataset was bootstrapped from extractions of the OpenIE4 system~\cite{OpenIE4}.
It contains n-ary extractions, enabling model evaluation on both binary and n-ary extraction benchmarks. 
We also randomly select 42k annotated sentences from the original training data and triggered the large language model (gpt-3.5-turbo) to obtain the labeled dataset based on our prompts. 
The dataset contains 5 languages: Arabic, Chinese, German, Portuguese, and Spanish. The new annotated dataset coupled with original English data aggregates the \dataset. Table \ref{tab:dataset_statistics} lists the statistics.


\item Re-OIE2016~\cite{SpanOIE} is a more accurate English n-ary extraction benchmark that is manually re-annotated the entire OIE2016.
Spanish and Portuguese versions of Re-OIE2016 are extended by~\cite{multiligual_OIE}, with the same number of sentences and tuples for each language.

\item CaRB~\cite{carb} is an English n-ary extraction benchmark which is a crowd-sourced re-annotated dataset based on dev and test splits of OIE2016. It has higher coverage and quality of the reference extractions compared to most of the OIE benchmarks.

\item BenchIE~\cite{BenchIE2021} is a multi-lingual OIE benchmark for binary extraction evaluation in English, Chinese, and German. 
Unlike other datasets, BenchIE is an exhaustive fact-based benchmark that includes fact synsets, Each synset is a set of all acceptable surface forms of the same fact.
In other words, the gold standard into account the informational equivalence of extractions, which makes evaluation more comprehensive.

\end{itemize}

\paragraph{Evaluation Metrics}
We evaluated each OIE system using the F1-score.
The F1-score is a balanced assessment of the model, combining precision and recall into a single measure. 
We used the evaluation code provided with each benchmark. allowing the extractions to be slightly different from the gold tuples, as there are no restrictions on the elements of open extractions. 


For the CaRB evaluation, we utilize their \textit{tuple match} which is a stricter token-level matching scorer for a rigorous evaluation. It matches predicted predicates with golden predicates and predicted arguments with golden arguments respectively.
 Since the \textit{lexical match} evaluation has numerous shortcomings~\cite{carb}, we also use \textit{tuple match} matching criterion on the multi-lingual Re-OIE2016.
BenchIE provides a fact-level matching scorer which takes the informational equivalence of extractions into account by exactly matching extracted triple with the corresponding gold fact synset (i.e., the same fact with different surface forms).

\paragraph{Implementation Details}
We train the model for 1 epoch in each stage. 
The batch size is set to 128 in the first
and second stages, and 64 in the third stage. 
The maximum sentence length is set to 100.
The number of experts in each \mlora{} is set to 6, and we tune the best LoRA rank is 64.
We use AdamW~\cite{adamw} as our optimizer with an
initial learning rate of 3e-5. For the cross-lingual encoder, we use the multi-lingual
BERT (mBERT) \cite{bert}.
The model is trained on a single NVIDIA Tesla V100 (32GB). We choose the top-$4$ LoRA experts based on the best average F1 score at the inference stage.

\subsection{Baselines}
We compare our model with both English and multi-lingual baselines. For the evaluation of the English datasets, we use non-neural systems:
Stanford~\cite{Stanford},
ClausIE~\cite{ClausIE},
MinIE~\cite{MinIE} and neural models:
RnnOIE~\cite{RnnOIE},
SpanOIE~\cite{SpanOIE},
IMoJIE~\cite{imojie},
CIGL~\cite{OpenIE6}
OpenIE6~\cite{OpenIE6},
Multi2OIE~\cite{multiligual_OIE}.

For the evaluation of multiple languages, Multi2OIE~\cite{multiligual_OIE} is used as neural-network-based baselines. Rule-based systems like ClausIE and MinIE cannot be used for languages other than English. We use
ArgOE~\cite{ArgOE} and PredPatt~\cite{PredPatt} as rule-based baselines, which are only two multi-lingual systems.  

\subsection{Main Results}
\subsubsection{English}\label{exp:english}
We compare our model with several unsupervised and supervised baselines on CaRB and BenchIE English benchmarks. 
Compare to rule-based and neural-based models, \ourmethod{} achieves a relatively high F1 score on CaRB n-ary extractions. 
Constrained-IGL (CIGL) is an individual component in OpenIE6, which achieves the highest performance among all prior models but can only use English-specific constraints in training.
The performance gap between \ourmethod{} and the multi-lingual baseline Multi2OIE~\footnote{The results reported in~\cite{multiligual_OIE} are based on the English version and we reproduce it by loading the officially released multi-lingual version checkpoint for a fair comparison in this study.} is minimal on CaRB n-ary extraction. 
Since CaRB's evaluation scheme penalizes long extraction in the precision calculation, however, it may cause high recall just simply adding words in extraction.
We notice that even though \ourmethod{} cannot reach the highest F1 score on CaRB, it yields the highest precision score, which is more convincing in this evaluation scheme.

\ourmethod{} performs best compared to other neural-based baselines on BenchIE binary extraction. Even though rule-based systems like ClausIE and MinIE outperform all neural systems, they cannot be used for non-English languages. Similar to the result on CaRB, the 
 high performance on BenchIE is attributed to the high precision.

\begin{table}[htb]

\centering
\resizebox{1.0\columnwidth}{!}{
\begin{tabular}{lcccccc}
\toprule
Models & \multicolumn{3}{c}{CaRB}  & \multicolumn{3}{c}{BenchIE} \\
\midrule
& F1 & PREC. & REC. & F1 & PREC. & REC. \\ 
ClausIE & 44.9 & - & - &33.9&50.3&25.6\\ 
MinIE & 41.9 & - & - &33.7&42.9&27.8 \\
Stanford & 23.9 & - &- &13.0 &11.1&15.7\\
\midrule
RnnOIE & 46.7 & 55.6 & 40.2 &13.0 & 37.3& 7.8\\
SpanOIE & 49.4 & 60.9 &41.6 & - &  -& - \\
IMoJIE & 53.5 & - & - & - &  -& - \\ 
CIGL & \textbf{54.0} & - & - & - &  -& - \\ 
OpenIE6 & 52.7 &- &- & 25.4 & 31.1& \textbf{21.4} \\
Multi2OIE & 51.9 & 59.5 & \textbf{45.9} & 23.8 & 37.7 & 17.4 \\
\midrule
\textbf{\ourmethod{}} & 51.8 & \textbf{65.8} & 42.7 & 29.1 & 50.0 & 20.5 \\

\textbf{- \mlora{}} & 51.6 & 65.6 & 42.5 & 28.6 & 48.8 & 20.0 \\

\textbf{- \dataset} & 51.3 & 64.9 & 42.4 & \textbf{29.3} & \textbf{50.5} & 20.7 \\

\bottomrule
\end{tabular}}
\caption{\ourmethod{} performance comparison with baseline models on CaRB n-ary and BenchIE binary English extraction benchmarks. 
}
\label{tab:carb&benchie}
\end{table}

\begin{table}[htb]
\centering
\resizebox{1.0\columnwidth}{!}{
\begin{tabular}{c|c|c|c|c}
\toprule
Language & System & F1 & PREC. & REC. \\
\midrule
\multirow{4}{*}{\centering En} 
   & ArgOE      & 43.4    & 56.6 & 35.2  \\
   & PredPatt   & 53.1    & 53.9 & 52.3  \\
   & Multi2OIE  & 69.3    & 66.9 & \textbf{71.7}  \\
   & \textbf{\ourmethod{}} & \textbf{69.5} & \textbf{73.4} & 66.0  \\
   \midrule
   \multirow{4}{*}{\centering Pt}
   & ArgOE      & 38.3 & 46.3 & 32.7  \\
   & PredPatt   & 42.9 & 43.6 & 42.3  \\
   & Multi2OIE  & 59.1 & 56.1 & \textbf{62.5}  \\
   & \textbf{\ourmethod{}} & \textbf{60.7} & \textbf{63.5} & 58.2  \\
\midrule
\multirow{4}{*}{\centering Es} 
   & ArgOE      & 39.4 & 48.0 & 33.4  \\
   & PredPatt   & 44.3 & 44.8 & 43.8  \\
   & Multi2OIE  & 60.2 & 59.1 & \textbf{61.2}  \\
   & \textbf{\ourmethod{}} & \textbf{61.0} & \textbf{65.0} & 57.5  \\
\bottomrule
\end{tabular}}
\caption{The models are tested using CaRB's evaluation scheme \textit{tuple match} for rigorous evaluation on the multi-lingual Re-OIE2016. The results are cited from~\cite{multiligual_OIE}, 
which only reported binary extraction performance due to the baseline systems being binary extractors. 
}\label{tab:reoie2016}
\end{table}



\begin{table*}[htb]
\centering
\resizebox{0.7\textwidth}{!}{
\begin{tabular}{l|ccc|ccc|ccc}
\toprule

\multirow{2}{*}{} & \multicolumn{3}{c|}{Zh} &
\multicolumn{3}{c|}{De} & \multicolumn{3}{c}{Ar}\\
  & F1 & PREC. & REC.  & F1 & PREC. & REC. & F1 & PREC. & REC.\\
\midrule

Multi2OIE  & 9.0 & 11.0 & 7.6 & 3.3 & 5.7 & 2.3 & 4.4 & 12.5 & 2.7 \\
\midrule
\textbf{\ourmethod{}}  & \bf 16.0 & \bf 23.1 & \bf 12.3 & 4.4 &  8.5 &  2.9 & \bf 11.0 & \bf 23.8 & \bf 7.2 \\

\textbf{- \mlora{}}  &  14.9 &  20.9 &  11.6 &  4.1 &  8.0 &  2.8 & 7.0 &  18.3 &  4.3 \\

\textbf{- \dataset}  &  14.9 &  21.1 &  11.6 & \bf 5.0 & \bf 11.6 & \bf 3.2 &  4.4 &  19.7 & 2.5 \\

\bottomrule
\end{tabular}}
\caption{The performance of multi-lingual neural-based OIE models on BenchIE-multi-lingual binary extraction. The results of Multi2OIE are reproduced in multi-lingual settings.}\label{tab:benchie}
\end{table*}

\subsubsection{Multi-lingual}\label{exp:multi-lingual}
In Table~\ref{tab:reoie2016}, we compare our model with multi-lingual baselines on Re-OIE2016 multi-lingual version benchmark that is proposed by~\cite{multiligual_OIE}. 
\ourmethod{} outperforms the other baselines in all languages and yields the highest F1 values, which has demonstrated the excellent cross-lingual abilities of our framework. Specifically, \ourmethod{} outperforms Multi2OIE by 0.2\%, 0.8\%, and 1.6\% in English, Spanish, and Portuguese, respectively. The superiority of our framework is attributed to its high precision, which is more reliable since CaRB evaluation rewards long extractions with much higher recall scores as we discussed in Section~\ref{exp:english}.

In Table~\ref{tab:benchie}, we compare \ourmethod{} with the multi-lingual neural-based model on the BenchIE non-English datasets. 
Similar to the method proposed in~\ref{sec:multi-lingual_training}, we triggered gpt-3.5-turbo to annotate 100 sentences from BenchIE-English data to Arabic. Then we amended all the incorrect triples manually with the help of a native Arabic speaker.
From Table~\ref{tab:benchie}, \ourmethod{} outperforms Multi2OIE in all languages. Chinese and Arabic have a significant improvement over the baseline model.
We can observe that both multi-lingual models perform significantly worse in German. There are many German verb stems and their separable prefixes appear in sentences. The decoding method of both models used a BIO tagging scheme that identifies continuous phrases, which are always absent in predicates, resulting in an extremely low recall.



\begin{figure}[t]
\begin{center}
	\includegraphics[width=1.0\columnwidth]{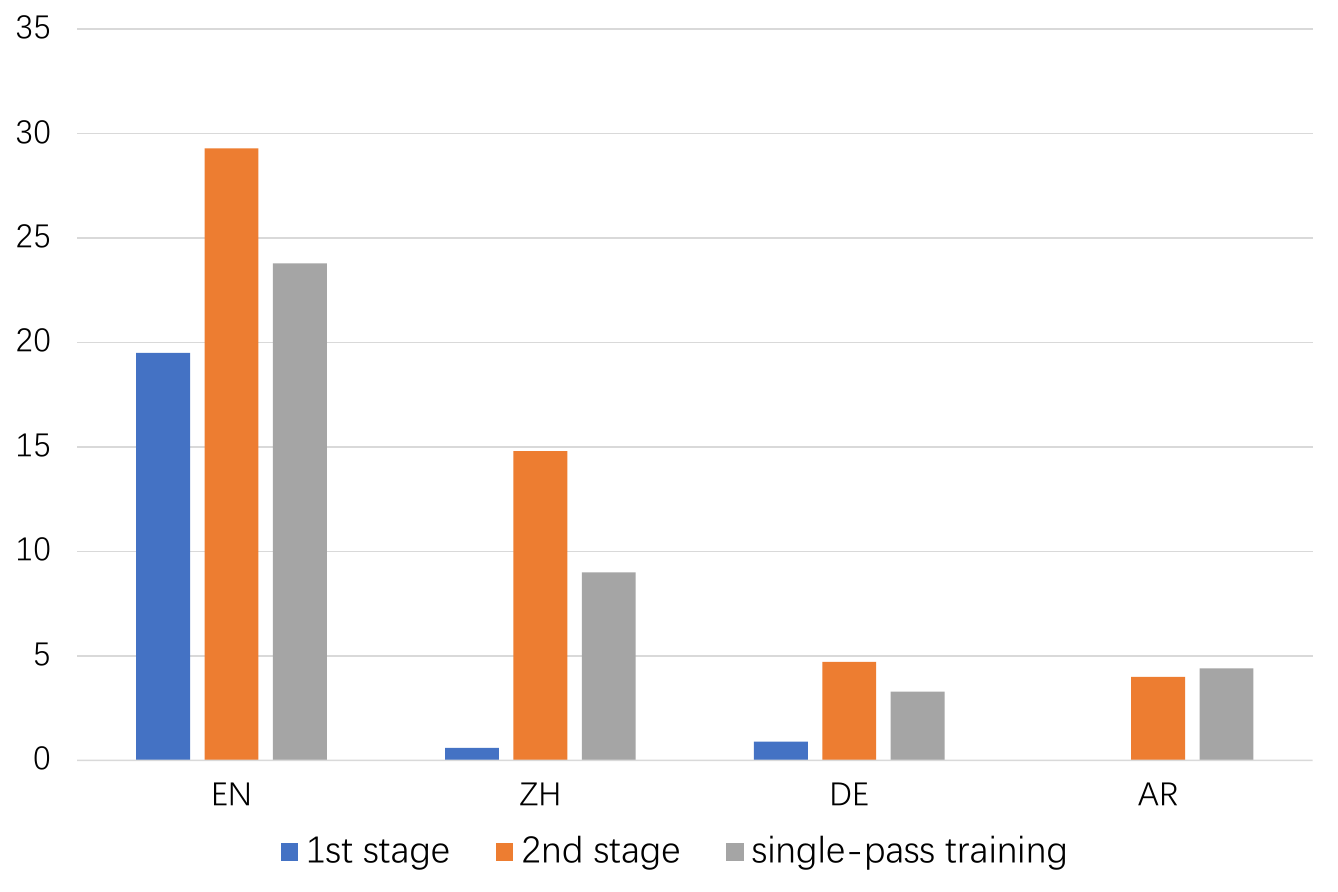}
	\caption{The comparison among different disentangled tuning stages and one pass training strategy.}
	\label{disentangled_comparison}
\end{center}
\end{figure}

\subsection{Effectiveness of Disentangled Tuning}
We use a high-quality multi-lingual dataset BenchIE to explore the effect of the disentangled tuning strategy.
From Figure~\ref{disentangled_comparison}, we observe that our model reaches higher performance on multiple languages after disentangled tuning compared to Multi2OIE which is tuned on a single pass. It is apparent that our framework truly keeps knowledge from being forgotten from the big picture. Since we tune the different parts in English training data, the language and task features are learned adequately in English even without prior knowledge in the first stage, while a big gap in the first and second stages of the other three languages. The results in Chinese and German demonstrate the model’s satisfactory zero-shot performance even though non-English data is not available in the training stages. However, the results in Arabic seem slightly lower, even with no performance in the first stage. We suppose that English, Chinese, and German are subject-verb-object languages, while Arabic is a verb-subject-object language, and their subjects or objects can be expressed as part of the verb, resulting in low performance. Such interference may hurt model performance in a certain language during our disentangled tuning. 

\subsection{Ablation Study}
To investigate how different parts influence the overall performance, we conduct an ablation study on the third stage. 
From Table~\ref{tab:carb&benchie} and Table~\ref{tab:benchie}, we observe that all components are helpful for the proposed method. In particular, there is an evident performance drop across all languages when removing the \mlora{} from our proposed method, which indicates the effectiveness of the \mlora{} for improving the capability of the model-based cross-lingual transfer. It also demonstrates that different experts can provide diverse knowledge to enrich the limited language-specific representation.

Moreover, without the help of \dataset, training \mlora{} only with the raw English corpus (OpenIE4) will cause a distinct performance drop in other languages, which demonstrates the effectiveness of the data-based cross-lingual transfer. 
This indicates that training with a mixture of multiple languages contributes to improving the low-resource language representation, especially in non-Indo-European language families (e.g., Chinese and Arabic). 
Interestingly, we observe that German has a small performance enhancement without a German training corpus.
We assume that the limited German training corpus did not provide much help due to their sophisticated language feature applied to the BIO scheme and the interference of the other languages can hurt the model performance on German.
Besides, English and German have many similarities as they are in the same language family, which also benefits the cross-lingual transfer.
We notice that performance on BenchIE English also has a minimal improvement compared to \ourmethod{}, while a slight drop on CaRB. We suppose the performance decline is mainly caused by model overfitting on CaRB. Our full model has achieved a great balance.

\begin{table*}[t]
\resizebox{1.0\textwidth}{!}{
\begin{tabular}{l|l}
\toprule
\toprule
\textbf{Sent. \#1} & \makecell[l]{Sligo town then became an incorporated municipal borough with a Royal Charter issued by the British King James I in 1613/14.}  \\ 
\midrule
Gold & \makecell[l]{[a] Royal Charter --> issued by --> [the] British King \\
{[a]} Royal Charter --> issued --> by [the] British King \\
{[a]} Royal Charter --> issued by --> [the] [British] [King] James I \\
{[a]} Royal Charter --> issued --> by [the] [British] [King] James I} \\ 
\midrule
Multi2OIE & \makecell[l]{a Royal Charter --> issued --> by the British King James I \colorbox{red!30}{in 1613/14}}  \\
\midrule
\ourmethod{} & \makecell[l]{a Royal Charter --> issued --> by the British King James}  \\ 
\midrule

\textbf{Sent. \#2} & \makecell[l]{It hosts the `` Zomercarnaval '', the second largest Caribbean carnival in Europe, originally called the Antillean carnival.} \\ 
\midrule
Gold &  \makecell[l]{It --> hosts --> [the] [``] Zomercarnaval ['']\\
It --> hosts --> [the] ``Zomercarnaval'' \\
It --> hosts --> [the] [second largest] Caribbean carnival [in Europe] \\
It --> hosts --> [the] Antillean carnival} \\ 
\midrule
Multi2OIE & \makecell[l]{It --> hosts --> the `` Zomercarnaval '' \colorbox{red!30}{the second largest Caribbean carnival in Europe}} \\ 
\midrule
\ourmethod{} & \makecell[l]{It --> hosts --> the `` Zomercarnaval} \\ 
\midrule

\textbf{Sent. \#3} &  \makecell[l]{The Anti-Monitor began to siphon the positive matter of New York City to create his Antimatter waves.} \\ 
\midrule
Gold &   \makecell[l]{
{[The]} Anti-Monitor --> [began to] siphon --> [the] positive matter [of New York City] \\
{[The]} Anti-Monitor --> began --> to siphon [the] positive matter [of New York City] \\
{[The]} Anti-Monitor --> began to --> siphon [the] positive matter [of New York City] \\
{[The]} Anti-Monitor --> began to siphon --> [the] positive matter [of New York City]} \\ 
\midrule
Multi2OIE & \makecell[l]{The Anti-Monitor --> began to siphon the positive matter of \colorbox{red!30}{New} --> \colorbox{red!30}{York City}} \\ 
\midrule
\ourmethod{} &  \makecell[l]{The Anti-Monitor --> began to siphon --> the positive matter of \colorbox{blue!30}{New York City}} \\ 
\midrule

\textbf{Sent. \#4} &  \makecell[l]{Salomon Brothers says, `` We believe the real estate properties would trade at a discount ... after the realty unit is spun off ... .}
\\ 
\midrule
Gold &  \makecell[l]{Salomon Brothers --> says --> [,] We believe [the] real estate properties would trade at [a] discount [...] after [the] realty unit is spun off [...] \\
Salomon Brothers --> says --> [,] We believe [the] real estate properties would trade at [a] discount \\
Salomon Brothers --> says --> [,] [``] We believe [the] real estate properties would trade at [a] discount [...] after [the] realty unit is spun off [...] \\
Salomon Brothers --> says --> [,] [``] We believe [the] real estate properties would trade at [a] discount \\
} 
\\ 
\midrule
Multi2OIE & \makecell[l]{Salomon Brothers --> says --> We believe the real estate properties would trade at a discount ... after the realty unit is \colorbox{red!30}{spun}} \\ 
\midrule
\ourmethod{} &  \makecell[l]{Salomon Brothers --> says --> We believe the real estate properties would trade at a discount ... after the realty unit is \colorbox{blue!30}{spun off}} \\ 
\bottomrule
\bottomrule
\end{tabular}}
\caption{Cases of baseline and our method. We only select four annotations that are most similar to the output as golden references. [$\cdot$] denotes an optional item. The two arguments and the corresponding predicate are separated by -->. Error parts are highlighted in red-colored font, while correct parts are in blue-colored.}
\label{tab:case_study}
\end{table*}

\subsection{Case Study}
To provide an in-depth analysis of the cases, we examine the extraction outputs of our proposed \ourmethod{} and the Multi2OIE baseline from several random samples on the BenchIE English benchmark, as shown in Table~\ref{tab:case_study}.
The following samples exemplify four advantages of \ourmethod{}. We summarize the four major superiorities of our framework in these samples:
\paragraph{Concise Extraction}As shown in Sent. \#1, our framework concisely extracts the target triplet, while the baseline model appends unnecessary date information at the end of the sentence.
\paragraph{Coreference Resolution}In Sent. \#2, our method is capable of identifying the coreference. Instead, the baseline model extracts an appositive clause, making the extraction redundant and confusing.
\paragraph{Named Entity Recognition}In Sent. \#3, the named entity ``New York City'' is correctly recognized by our method. However, the baseline model fails to recognize it as a whole.
\paragraph{Grammatical Correctness}In Sent. \#4, the preposition ``off'' is missing in the baseline model extraction, causing a grammatical error, while our framework adds it accurately.



\subsection{Discussion}
\paragraph{The Effect of LoRA Rank}
\begin{table*}[htb] 
\centering
\resizebox{0.8\textwidth}{!}{
\begin{tabular}{l|c|cccc|ccc|c}
\toprule

\multirow{2}{*}{} & \multicolumn{1}{c|}{CaRB} & \multicolumn{4}{c|}{BenchIE} & \multicolumn{3}{c}{Re-OIE2016}& \multicolumn{1}{|c}{\centering \textbf{Total}}\\

 & \multicolumn{1}{c|}{En} & \multicolumn{1}{c}{En} & \multicolumn{1}{c}{Zh} & \multicolumn{1}{c}{De} & 
 \multicolumn{1}{c|}{Ar} & 
 \multicolumn{1}{c}{En} & \multicolumn{1}{c}{Pt} & \multicolumn{1}{c}{Es} & \multicolumn{1}{|c}{}\\
\midrule
rank = 1 ($k$=1) & 51.6 & 29.2 & \bf 17.1 & 3.9 & 9.8 & 69.4 & 61.2 & 60.6 & 302.8 \\
rank = 2 ($k$=3)  & 51.4 & \bf 29.3 & 16.6 & 3.7 & 10.5  &  69.4 & 61.0 & 61.1 & 303.1 \\
rank = 4 ($k$=3)  & 51.5 & 29.0  & 16.2 & 3.6 & 9.8 & 69.3 & \bf 61.3 & 60.9 & 301.6 \\
rank = 8 ($k$=6) & 51.5 & 29.2 & 16.4 & 3.7 & 9.4 & 69.4  & 61.0 & 61.1 & 301.9 \\
rank = 16 ($k$=6)   & 51.6 & 28.8 & 16.7 & 4.0 & 9.8 & \bf 69.5 & 61.1  & \bf 61.3 & 302.6 \\
rank = 32 ($k$=2) & 51.7 & 29.1 & 15.7 & 4.1 & \bf 11.4 & 69.2  & 61.1 & 61.3 & \bf 303.5 \\
rank = 64 ($k$=4, ours)  & \bf 51.8 & 29.1 & 16.0  & \bf 4.4 & 11.0 & \bf 69.5 & 60.7 & 61.0 & \bf 303.5 \\
rank = 128 ($k$=2) & \bf 51.8 & 28.8 & 16.4 & \bf 4.4 & 10.5 & 69.4 & 60.5 & 60.9 & 302.6 \\
\bottomrule
\end{tabular}}
\caption{Performance of \ourmethod{} with different LoRA ranks on benchmarks. We choose the best top-$k$ value for each rank. The bold font \textbf{Total} denotes the sum of F1 scores across all datasets.}
\label{tab:lora_rank}
\end{table*}
Given a limited memory budget, what is the optimal combination of rank $r$ for our \mlora{} module in the top-$k$ strategy? The results of the effect of $r$ on model performance are presented in Table~\ref{tab:lora_rank}.
Note that \mlora{} performs competitively with a relatively big $r$ (e.g., rank=32 and rank=64).
We argue that increasing $r$ to a certain degree does cover a more meaningful subspace. We notice that rank=2 also achieves a considerable performance. This is more desirable if considering the model capacity.

\paragraph{Cross-lingual Representation of Multi-stage Training}
\begin{figure}[t]
\centering
    \subfigure[Pretrained Model]{
    \includegraphics[width=0.43\columnwidth]{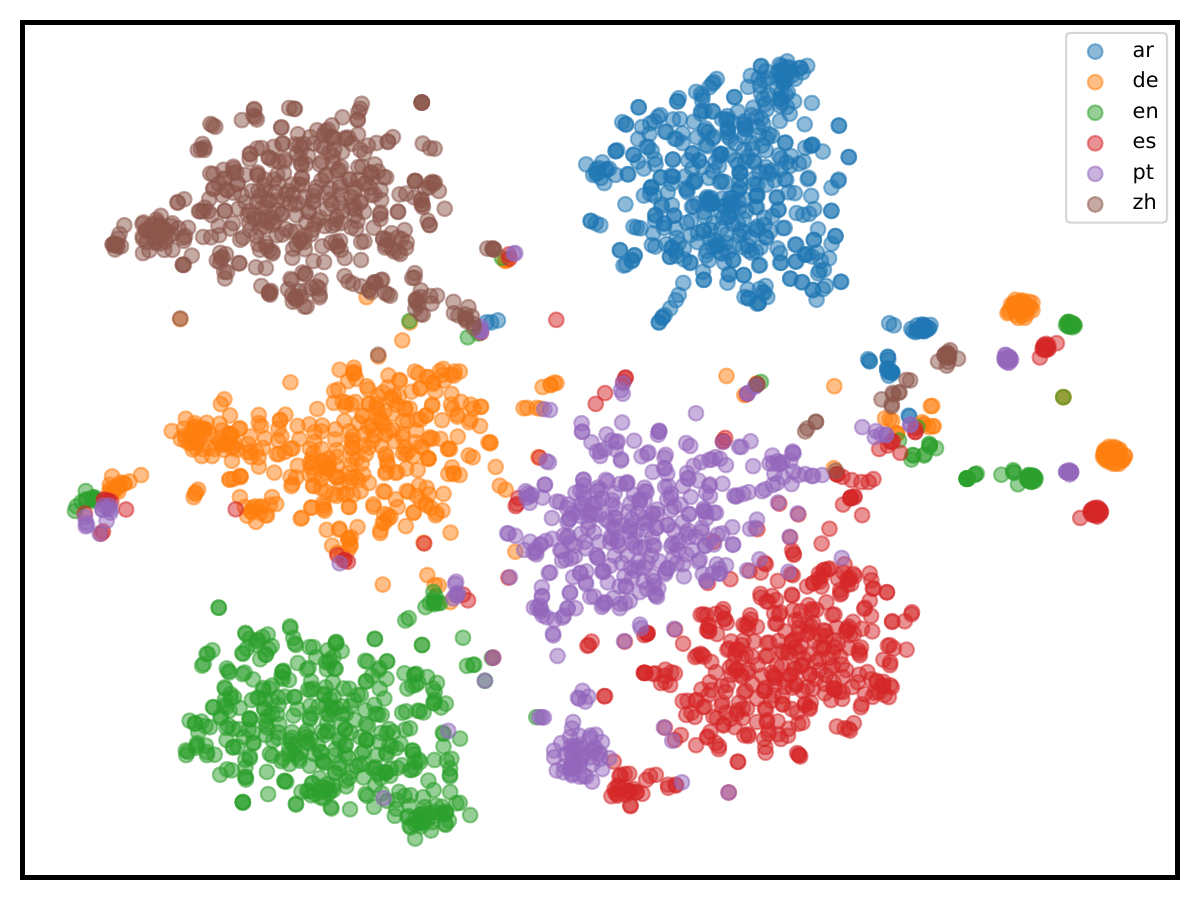}
    \label{tsne_0stage}
    }
    \subfigure[First Stage]{
    \includegraphics[width=0.43\columnwidth]{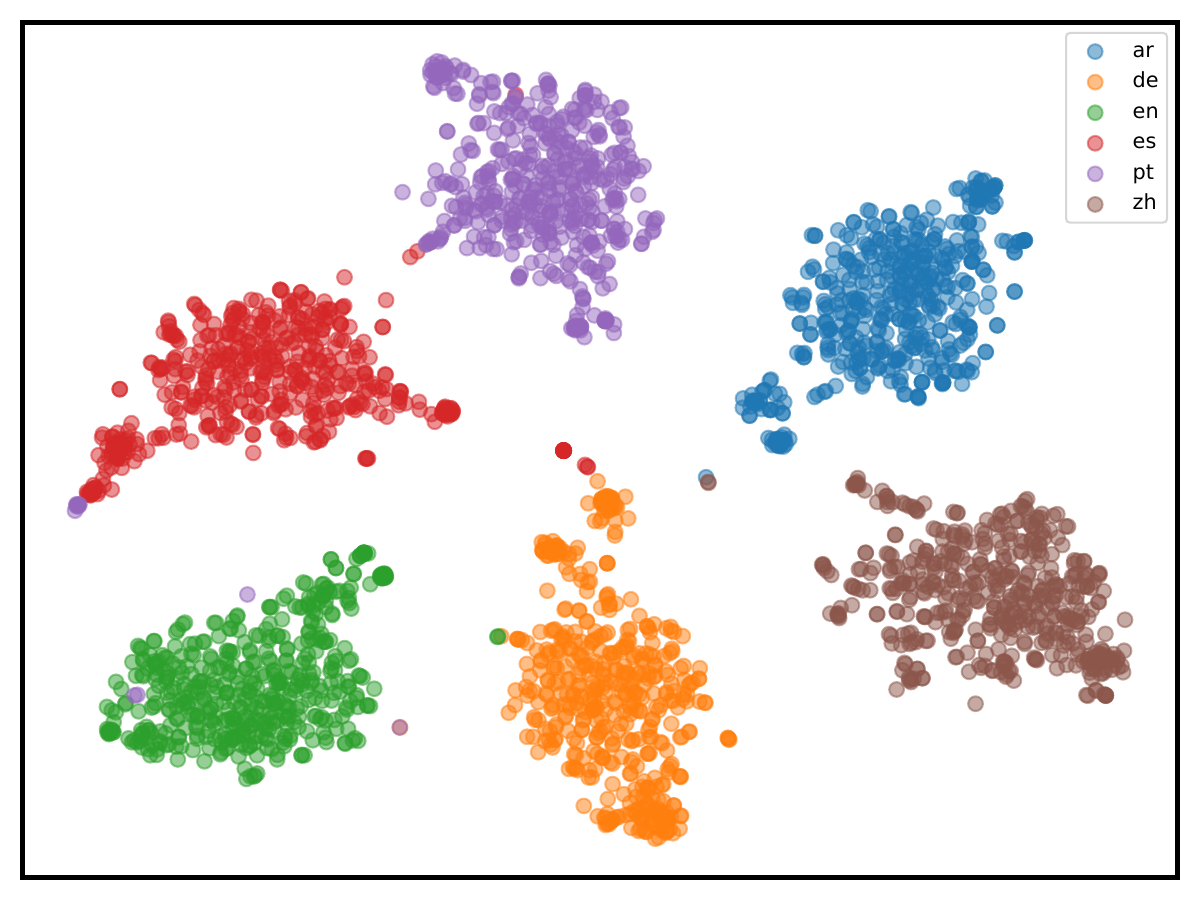}\quad
    \label{tsne_1stage}
    }
    \subfigure[Second Stage]{
    \includegraphics[width=0.43\columnwidth]{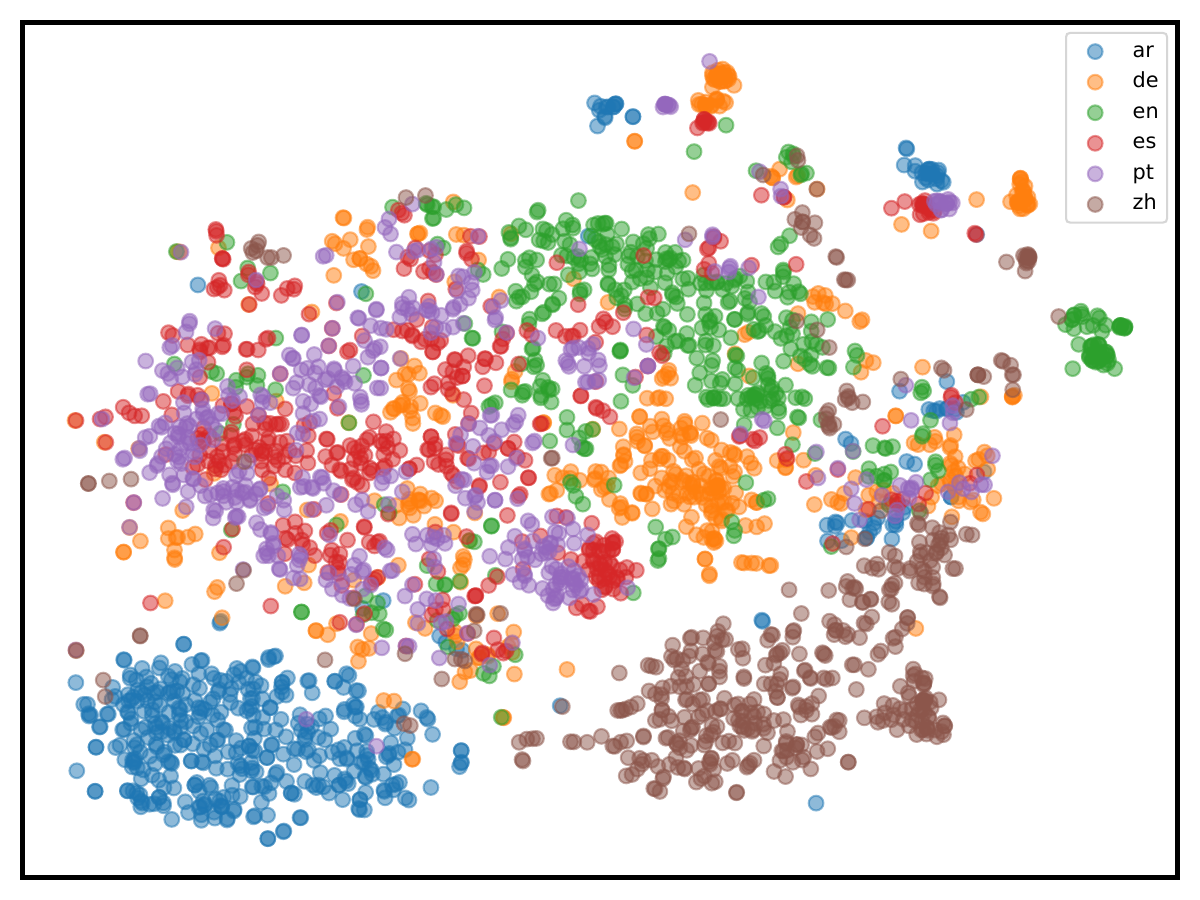}
    \label{tsne_2stage}
    }
    \subfigure[Third Stage]{
    \includegraphics[width=0.43\columnwidth]{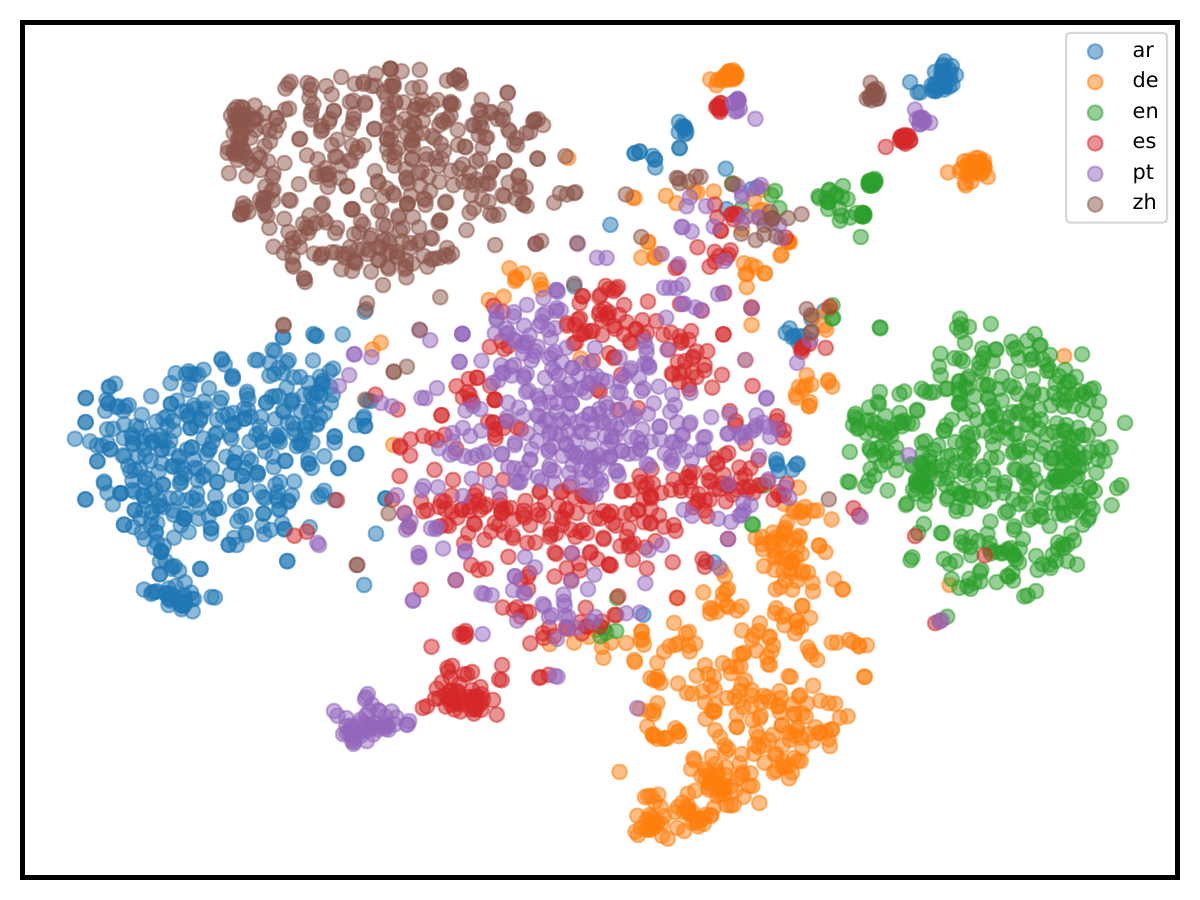}\quad
    \label{tsne_3stage}
    }
    \caption{t-SNE \cite{t_SNE} visualization of the average sentence representations in the \dataset{} dataset for multi-stage training strategy. (a) are initial representations of cross-lingual pre-trained models. (b) are features after the first stage. (c) are features after the second stage. (d) are features after the third stage. } 
    \vspace{0pt}
    \label{tsne_figures}
\end{figure}
(a) We load pre-trained parameters of mBERT without any fine-tuning to observe the sentence representations directly across different languages. The language representations are scattered in space after prior pre-training in 104 languages.
(b) From the observation of our first stage, the language representations have become even more scattered. There is an evident distinction among languages. 
(c) After tuning in the second stage, all language representations are mixed together. The languages are aligned in a shared space, where the similar semantic representations of languages are close in the same position area. The model obtains the benefits of shared parameters after the disentangled tuning.
(d) We observe that the language representations are slightly scattered again in the third stage. That indicates the languages perceive \mlora{} in the third stage, which means language features are well-distinguished after obtaining independent parameters while retaining the benefits of shared parameters.

\section{Related Work}

\paragraph{Open Information Extraction} Open Information Extraction (OIE) is a task that extracts a set of n-ary relation tuples from an arbitrary domain text~\cite{survey_oie}. OIE systems have two main categories: (I) Unsupervised rule-based approaches, which perform extractions with dependency parsers and PoS taggers based on fine-grained rules or handcrafted features~\cite{guo2023loglg,Fader2011,OLLIE,ClausIE,MinIE2017,MinScIE2019}. Most recent OIE approaches are usually based on neural networks~\cite{liu2020block} which are built as different supervised learning models. Neural solutions become popular and achieved considerable improvement due to the large-scale OIE benchmarks~\cite{StanovskyD16,carb,SpanOIE}.
(II) Supervised neural OIE models, which handle OIE tasks by utilizing sequence labeling models to tag each token as a role label in a sentence~\cite{StanovskyD16,StanovskyMZD18,RoyPLP19,SarhanS19,multiligual_OIE,JiaSDCX22}, using span-based models to directly predict whether a span-level phrase is a predicate or an argument instead of a BIO tag in a token-level~\cite{SpanOIE}, or performing an encode-decode schema to produce extraction tuples as a sequence step by step using sequence generation models~\cite{CuiWZ18,KolluruARMC20}.
In this paper, we view OIE as a sequence labeling problem and build up \ourmethod{}. 

\paragraph{Cross-lingual NLP Tasks} Cross-lingual tasks include various NLP tasks involved in multiple languages~\cite{bert,mbart,xnlg,liu2022cross,guo2022lvp}, such as cross-lingual pre-training \cite{xlm,xlmr,alm,ganlm}, cross-lingual named entity recognition \cite{con_ner,crop}, and cross-lingual summarization \cite{cross_sum}, and multi-lingual translation \cite{cluster_mnmt,hlt_mt,um4}. Cross-lingual transfer (the process of leveraging knowledge and resources from one language to another) plays a pivotal role in cross-lingual tasks. This approach not only saves resources but also helps overcome the data scarcity problem in low-resource languages. Most of the previous studies \cite{multiligual_OIE} can not be easily extended to the cross-lingual scenario of the OIE task, and thus our method is proposed to leverage the multi-stage training gradually distill the source language knowledge to other languages.

\section{Conclusion}

In this paper, we propose \ourmethod{}, a multistage tuning framework for cross-lingual open
information extraction, which injects language-specific knowledge into the shared model.
Moreover, we devise a novel data augmentation strategy, which leverages the chain-of-thought prompt to encourage the large language model annotating the multi-lingual raw data for data-based cross-lingual transfer.
Experimental results demonstrate that our approach outperforms the previous state-of-the-art approaches by a significant margin. Further analysis demonstrates our model effectively obtains language-agnostic representations in the shared parameters and language-specific knowledge in the mixture-of-LoRAs to reduce the gap among different languages.



\bibliography{custom.bib}
\bibliographystyle{acl_natbib}

\clearpage

\end{document}